\pdfoutput=1
\documentclass[11pt]{article}

\usepackage[preprint]{acl}

\usepackage{times}
\usepackage{latexsym}
\usepackage{graphicx}
\usepackage{amssymb}
\usepackage{makecell}
\usepackage{amsmath, bm}
\usepackage{longtable}
\usepackage{subfig}
\usepackage{caption}
\usepackage{float}

\usepackage{multirow}
\usepackage{enumitem}
\usepackage{algorithm}
\usepackage{algpseudocode}
\usepackage{booktabs}

\usepackage[T1]{fontenc}
\usepackage[utf8]{inputenc}

\usepackage{microtype}

\usepackage{inconsolata}

\usepackage{graphicx}

\title{Few-shot Named Entity Recognition via Superposition Concept Discrimination}

\author{
  Jiawei Chen${}^{1,3,}$,
  Hongyu Lin${}^{1}$\thanks{~ Corresponding authors.},
  Xianpei Han${}^{1,2}$,
  Yaojie Lu${}^{1}$,
  \\
  {\bf Shanshan Jiang$^{4}$}
  {\bf Bin Dong$^{4}$}
  {\bf Le Sun$^{1,2}$}
  \\
  ${}^{1}$Chinese Information Processing Laboratory ~
  ${}^{2}$State Key Laboratory of Computer Science \\
  Institute of Software, Chinese Academy of Sciences, Beijing, China\\
  ${}^{3}$University of Chinese Academy of Sciences, Beijing, China \\
  ${}^{4}$Ricoh Software Research Center Beijing Co., Ltd \\
  {\tt \{jiawei2020, hongyu, xianpei, yaojie, sunle\}@iscas.ac.cn} \\
  {\tt \{shanshan.jiang, bin.dong\}@cn.ricoh.com} \\
}

\begin{document}
\maketitle
\begin{abstract}
Few-shot NER aims to identify entities of target types with only limited number of illustrative instances. Unfortunately, few-shot NER is severely challenged by the intrinsic precise generalization problem, i.e., it is hard to accurately determine the desired target type due to the ambiguity stemming from information deficiency. In this paper, we propose Superposition Concept Discriminator (SuperCD), which resolves the above challenge 
via an active learning paradigm. Specifically, a concept extractor is first introduced to identify superposition concepts from illustrative instances, with each concept corresponding to a possible generalization boundary. Then a superposition instance retriever is applied to retrieve corresponding instances of these superposition concepts from large-scale text corpus. Finally, annotators are asked to annotate the retrieved instances and these annotated instances together with original illustrative instances are used to learn FS-NER models. To this end, we learn a universal concept extractor and superposition instance retriever using a large-scale openly available knowledge bases. Experiments show that SuperCD can effectively identify superposition concepts from illustrative instances, retrieve superposition instances from large-scale corpus, and significantly improve the few-shot NER performance with minimal additional efforts.
\end{abstract}

\section{Introduction}
Few-shot named entity recognition (FS-NER) aims to detect and classify named entity from text with only a few illustrative instances. 
FS-NER is appealing for open-domain NER which contains various unforeseen types and very limited examples, and therefore has attached great attention in recent years~\citep{fritzler2019few,yang-katiyar-2020-simple,ding-etal-2021-nerd,huang-etal-2021-shot}.

\begin{figure}[t!]
\centering 
\setlength{\belowcaptionskip}{-0.6cm}
\includegraphics[width=0.45\textwidth]{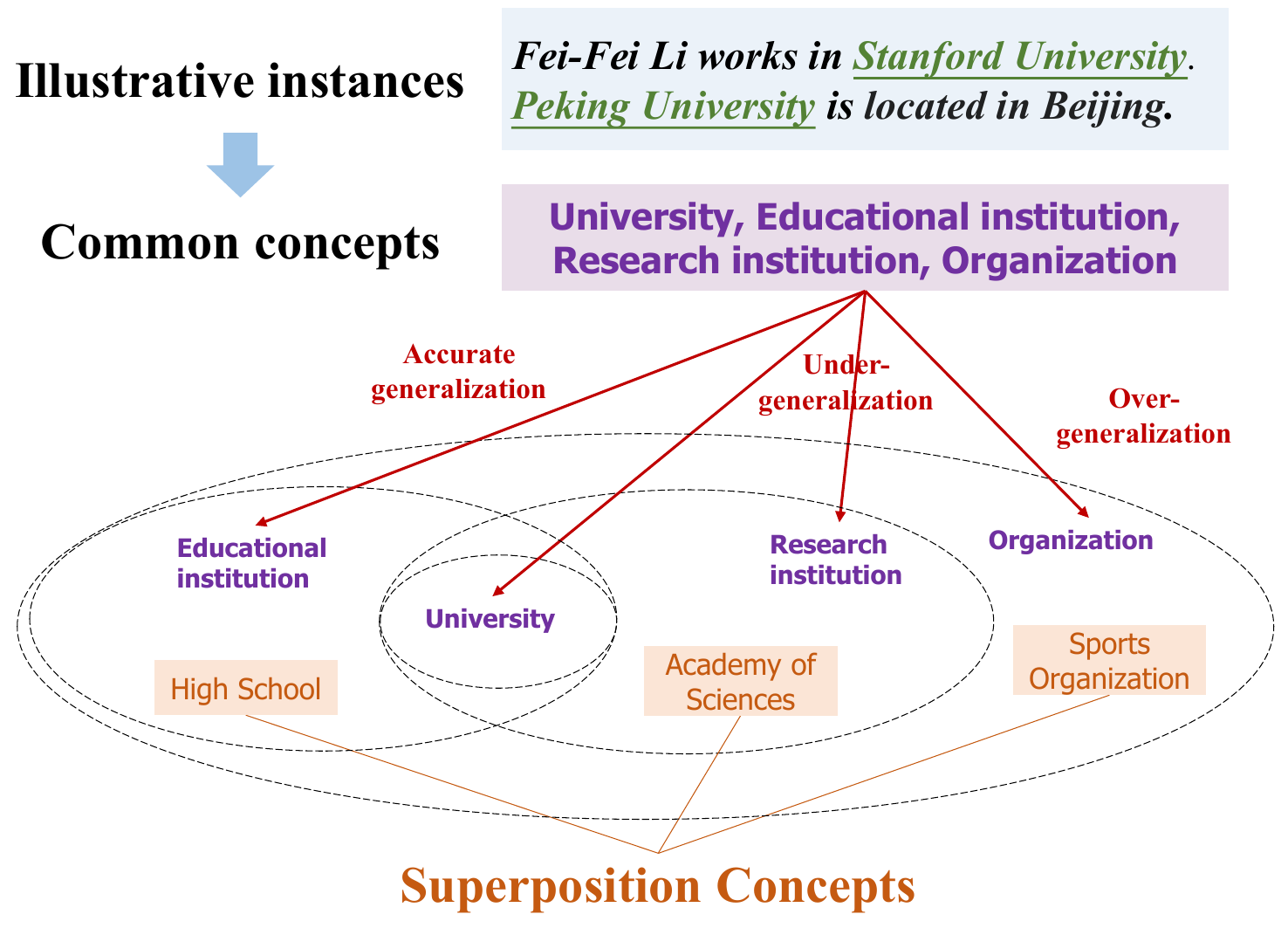}
\caption{Examples of the precise generalization challenge. The underline indicates the annotated entity mentions. Given only the illustrative instances, the desired target type may be \emph{University}, \emph{Educational institution}, \emph{Research institution} or \emph{Organization}. Discriminating superposition concepts like \emph{High school}, \emph{Academy of sciences} and \emph{Sports organization} helps determine what the desirable target entity type is.}
\label{Fig:exam}
\end{figure}

Even with rapid progress, FS-NER faces severe intrinsic \textit{\textbf{precise generalization}} challenge which is ignored by previous literature. Given only a few illustrative instances, it is frequently impossible to accurately determine what the desirable target entity type is. As a result, the learned FS-NER models often suffer from over-generalization or under-generalization. Figure~\ref{Fig:exam} shows an illustrative example. Given two illustrative instances ``Fei-Fei Li works in [Stanford University].'' and ``[Peking University] is located in Beijing.'', we are unable to determine the target entity type is \emph{University}, \emph{EduIns}(\emph{Educational institution}), \emph{ResIns}(\emph{Research institution}) or \emph{Organization}. Consequently, if the target entity type is \emph{EduIns}, a learned FS-NER model may unpredictably over-generalize to \emph{Organization} or \emph{ResIns}, it may also under-generalize to \emph{University}. Note that precise generalization is a task-intrinsic challenge of FS-NER because it can not be addressed by designing better model architecture without introducing additional accurate generalization knowledge.

\begin{figure*}[t!]
\centering 
\includegraphics[width=0.95\textwidth]{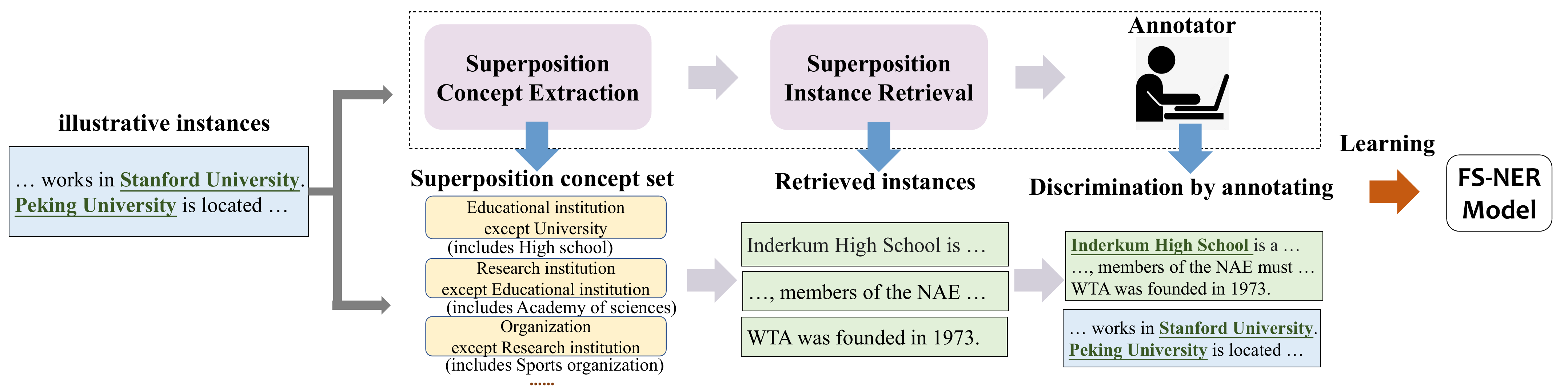}
\caption{Overview of SuperCD. The underline indicates the annotated entity mentions. SuperCD first extract sets of superposition concepts and then retrieve corresponding instances. Finally, by annotating that Inderkum High School is the target type while NAE (\emph{Research institution}) and WTA (Sports organization) is not and using these instances to learn model, the generalization knowledge is injected into the FS-NER model.}
\label{Fig:model}
\end{figure*}

The key to resolving the precise generalization challenge is to provide information about critical \emph{\textbf{superposition concepts}}, which refers to the concepts that are associated with some common concepts entailed in illustrative instances.
Without explicit declaration, instances of superposition concepts can both be or not be regarded as congener of illustrative instances. As a result, the target entity type can not be clarified unless providing sufficient measurement on these superposition concepts. For the example in Figure~\ref{Fig:exam}, \emph{HSchool} (\emph{High school}), \emph{AcademySci} (\emph{Academy of sciences}) and \emph{SportsOrg} (\emph{Sports organization}) are superposition concepts. On one hand, if we know that \emph{HSchool} is part of the desirable type, we can avoid the under-generalization because we know that \emph{University} is not the accurate target type. On the other hand, if we know that \emph{AcademySci} and \emph{SportsOrg} are not a desirable concepts, we can avoid over-generalizing to \emph{ResIns} and \emph{Organization}. As a result, a precise \emph{EduIns} recognizer can be learned only by providing additional information about \emph{HSchool}, \emph{AcademySci} and \emph{SportsOrg}. Unfortunately, accurately providing information about superposition concepts is very challenging due to the limited annotation budget and the unpredictable scale of potential superposition concepts. Consequently, simply annotating more instances is not only expensive but can not guarantee that the additionally-annotated instances can cover the superposition concepts that need to be measured. Therefore, how to identify critical superposition concepts and providing the information to FS-NER models with minimal additional efforts poses a huge challenge to FS-NER.\footnote{Only minimal additional instances (based on the number of types) will be provided, and therefore, we call our work few-shot NER.}

In this paper, we propose Superposition Concept Discriminator (SuperCD), which resolves the above-mentioned challenge in an active learning~\citep{lewis1994heterogeneous,settles2009active,shen2018deep,zhou2021mtaal} paradigm. Figure~\ref{Fig:model} shows the overall framework of SuperCD. The main idea behind SuperCD is to introduce an elimination-based approach to identify sets of superposition concepts and leverage an instance retriever to find instances corresponding to superposition concepts from a large-scale text corpus. Annotators are then asked to annotate the retrieved instances and these annotated instances and original illustrative instances are used to learn FS-NER models. In this way, the high-value precise generalization knowledge entailed in the additional annotated instances can be injected into the FS-NER models.
Specifically, to accurately identify the superposition concepts, we introduce a Concept Extractor (CE), and to retrieve instances of superposition concepts that need to be annotated, we introduce a Superposition Instance Retriever (SIR). Specifically, given a few illustrative instances of the target entity type, CE is first applied to extract the common concepts entailed in the illustrative instances. Then sets of superposition concepts are obtained based on an ``A but not B'' manner, in which the concept is part of concept A but not concept B. For the example in Figure~\ref{Fig:exam}, common concepts \emph{University}, \emph{Educational institution}, \emph{Research institution} and \emph{Organization} are first extracted, and then superposition concepts like \emph{``University but not Educational institution''}(which includes \emph{High school}), \emph{``Research institution but not Educational institution''} (which includes \emph{Academy of sciences}), \emph{``Organization but not Educational institution''}(which includes \emph{Sports organization}) and \emph{``University but not Organization''} will be constructed. After that, SIR will retrieve a certain number of instances of superposition concepts to be annotated from a large-scale text corpus based on the budgets. To equip CE with the ability of extracting universal concepts and SIR with the ability of instance retrieval, we learn CE and SIR on large-scale, easily accessible web resources~\citep{chen-etal-2022-shot}, which contains 56M sentences with more than 31K concepts from Wikipedia and Wikidata.

We conduct experiments on 5 few-shot NER benchmarks with different granularity. Experiments show that SuperCD significantly outperforms baselines and other active learning approaches under the same annotation budgets. Furthermore, SuperCD can effectively identify and discriminate superposition concepts. These demonstrate the effectiveness of SuperCD.\footnote{The code this paper: \url{https://github.com/chen700564/supercd}.}

Generally speaking, the contributions of this paper are:
\begin{itemize}
    \item We identify the precise generalization challenge which is a task-intrinsic of few-shot NER and is ignored by previous literature.
    \item We propose to resolve the precise generalization challenge by discriminating superposition concepts and propose an ``A but not B'' manner to identify superposition concept set.
    \item We propose Superposition Concept Discriminator (SuperCD), an active learning framework that injects generalization knowledge to FS-NER models by requiring annotators to provide a minimal number of additional annotated instances of superposition concepts.
\end{itemize}

\section{Related work}
\paragraph{Few-shot NER.} Previous works of FS-NER focused on better learning strategy and model architecture. Metric-based methods\citep{proto} are common on many FS-NER benchmarks with different structures like prototype network~\citep{fritzler2019few,tong-etal-2021-learning,wang-etal-2022-enhanced,ji-etal-2022-shot,wang-etal-2022-spanproto} and nearest neighbor network~\citep{yang-katiyar-2020-simple}. Prompt-based methods are promising for FS-NER~\citep{cui-etal-2021-template,liu2022qaner,ma-etal-2022-template}, which fully exploit the knowledge of pre-trained language models. Different learning strategies like contrastive learning\citep{das-etal-2022-container}, meta learning~\citep{li2020metaner,li2020few,de-lichy-etal-2021-meta,ma-etal-2022-decomposed} and self-training~\citep{huang-etal-2021-shot,wang2021meta} have been used to improve FS-NER. Recently, label semantics like type name and description is proved to be effective for FS-NER~\citep{DBLP:conf/acl/HouCLZLLL20,wang-etal-2021-learning-language-description,ma-etal-2022-label,chen-etal-2022-shot,yang-etal-2022-see,lee-etal-2022-good}, but it is difficult to obtain the description or type name accurately. Different from previous works, we focus on the task-intrinsic precise generalization challenge of FS-NER which cannot be addressed by designing better learning strategy or model architecture without introducing more generalization knowledge.

\paragraph{Active learning NER.} For NER, most of the previous active learning methods are uncertainty sampling~\citep{shen2018deep,huang2018low,agrawal2021active,DBLP:journals/corr/abs-2111-03837,zhou2021mtaal,liu2022ltp}, i.e., selecting the instances to be annotated based on the uncertainty scores. In addition, other related work considers data selection by multi-criteria~\citep{shen-etal-2004-multi,kim-2020-deep,nguyen-etal-2022-famie},annotation cost~\citep{10.1093/jamia/ocz102} etc. Different from previous works, we focus on few-shot scenarios with extremely limited budgets. Furthermore, the proposed SuperCD aims to introduce the generalization knowledge to FS-NER models by annotating minimal instances of superposition concepts for model training, which is different from previous works like uncertainty sampling methods.

\section{Superposition Concept Discriminator}
As we mentioned above, it is impossible to tackle the precise generalization challenge based only on the knowledge entailing in given few-shot instances. As a result, it is necessary to introduce additional instances to measure superposition concepts. Therefore, the main challenge here is how to identify superposition concepts and produce high-value instances for sufficient supervision with minimum annotation efforts.
To this end, we propose Superposition Concept Discriminator (SuperCD), an active learning-based framework for FS-NER. The overall architecture of SuperCD is shown in Figure~\ref{Fig:model}.  Specifically, given a few illustrative instances of the novel type, SuperCD first uses a concept extractor to extract the common concepts entailed in the illustrative instances. Then superposition concept sets are constructed based on an ``A but not B'' manner. After that, a Superposition Instance Retriever is applied to identify instances of each superposition concepts sets from large-scale raw corpus. Finally, annotators are asked to annotate superposition instances according to budget and all annotated and illustrative instances are used to learn few-shot NER models. In the following, we will first describe some critical concepts, and then illustrate each component of SuperCD.

\subsection{Definition of Superposition Concept}
In this paper, we use the term \textit{\textbf{concept}} to refer to a specific generalization of entity mentions, e.g., \emph{city}, \emph{human settlement}, \emph{location}, etc. Note that concepts here are universal, which may be or not be an entity type of a specific FS-NER task. In this paper, We collect 30k concepts from objects of ``Instance-of'' and ``Subclass-of'' relations in Wikidata.

We use the term \textit{\textbf{superposition concept}} to refer to the concept that is associated with some common concepts entailed in illustrative instances. 
Without additional information provided, an instance of superposition concept can both be or not be regarded as congener of illustrative instances, and therefore the target entity type can not be clarified unless providing sufficient measurement on these superposition concepts. 
For the example in Figure~\ref{Fig:exam}, \emph{High school} serves as a superposition concept because discriminating them is necessary to determine whether the target type is \emph{University} or \emph{Educational institution}. Unfortunately, due to the large scale of concepts and their compositions, it is very difficult to accurately recognize all superposition concepts for each FS-NER task. But in the following section, we will describe how we obtain the set of superposition concepts using the proposed concept extractor of SuperCD.

Without additional information, it is unclear whether an instance of the superposition concept should be considered as a congener of illustrative instances. Therefore, the target entity type cannot be determined without sufficient measurement on these superposition concepts. In Figure~\ref{Fig:exam}, for example, \emph{High school} serves as a superposition concept because distinguishing it is necessary to determine whether the target type is \emph{University} or \emph{Educational institution}. Unfortunately, accurately recognizing all superposition concepts for each FS-NER task can be challenging due to the large scale of concepts and their compositions. However, in the following section we will describe how we obtain the set of superposition concepts in SuperCD.

\subsection{Superposition Concept Extraction}\label{sec:te}
As we mentioned above, the main challenge for precise few-shot NER is how to identify the superposition concepts and how to provide additional knowledge about these superposition concepts to FS-NER models.

\begin{figure}[t!]
\centering 
\includegraphics[width=0.45\textwidth]{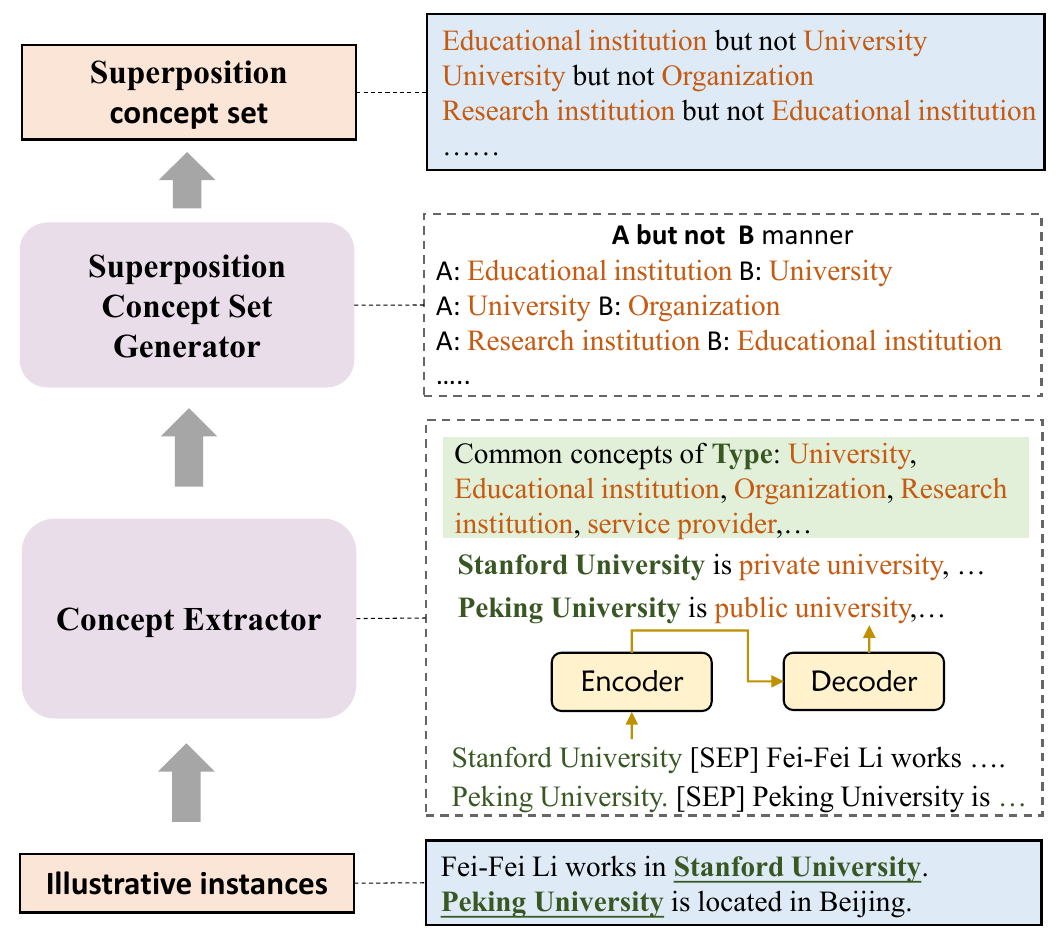}
\caption{The process of superposition concept extraction. Sets of superposition concepts are constructed through the concepts of few-shot illustrative instances with the ``A but not B'' manner.}
\label{Fig:sce}
\end{figure}

Unfortunately, it is impractical to directly recognize all superposition concepts due to the large amount of concepts in real world.
In this paper, instead of directly identifying each superposition concept, we propose to construct sets of superposition concepts using an elimination-based method. Figure~\ref{Fig:sce} shows an illustration of the entire superposition concept extraction procedure. Specifically, we first introduce a concept extractor, which can generalize illustrative instances into their common concepts. Then the superposition concepts are covered in the different sets of these common concepts. 

Formally, given a few illustrative instances $x_1,x_2,...,x_n$, the concept extractor will first generalize each instance $x_i$ into its corresponding universal concepts using a sequence-to-sequence concept extraction model~\cite{chen-etal-2022-shot}. The model takes illustrative instances $x_i$ with an annotated entity mention as input, and outputs the corresponding universal concepts:
\begin{equation}
C_i={\rm CE}(x_i)
\end{equation}
where $C_i = \{c_{i1},c_{i2},...,c_{ik}\}$ is the possible generalized concepts of illustrative instance $x_i$. Then we collect concepts generated from all illustrative instances, and regards concepts with high appearing frequencies as the common concepts of these few-shot instances. For example, given two illustrative instances in Figure~\ref{Fig:exam}, concept extractor will extract \textit{University}, \textit{Educational institution}, \textit{Research institution} as the common concepts. We use $C^{*} = \{c^{*}_{1},c^{*}_{2},...,c^{*}_{m}\}$ to represents the extracted common concepts. 

After that, we obtain sets of superposition concepts using an elimination-based method in an ``A but not B'' manner. Specifically, sets of superposition concepts are represented as ``$c^{*}_{i}$ but not $c^{*}_{j}$''. For example, given the common concepts \{\textit{University}, \textit{Educational institution}, \textit{Research institution},  \textit{Organization}\}, superposition concept sets like \emph{``Educational institution but not Organization''}, \emph{``Organization but not University''} and \emph{``Educational institution but not University''} will be constructed, and the superposition concepts \emph{High school}, \emph{Academy of sciences} and \emph{Sports organization} can be covered by the former sets respectively. In this way, we can obtain the superposition concept sets without knowing the exact superposition concepts.

\subsection{Superposition Instance Retrieval}\label{sec:ber}
Given sets of superposition concepts, another challenge to resolve the precise generalization issue is how to discriminate them and inject the information into FS-NER models.
One promising approach is to annotate a minimal number of instances of these superposition concepts, and directly uses them to learn FS-NER models. Therefore, it is necessary to accurately retrieve instances corresponding to superposition concepts from a large-scale corpus.

To this end, we design Superposition Instance Retriever (SIR), a dense retrieval architecture which regards the utterance of sets of superposition concepts as query, and retrieves texts containing such instances from the corpus. Formally, given a text piece $x$ in large-scale corpus and all superposition concept sets of the target entity type, we first construct superposition concept set queries by combining all sets with the same excluded concept. That is, the queries are in the form of $q =$ ``$c^{*}_{k} | c^{*}_{1},...,c^{*}_{k-1},c^{*}_{k+1},...,c^{*}_{m}$'', which represents that we want the retrieve instances of concept sets like ``$c^{*}_{1}$ but not $c^{*}_{k}$'', ``$c^{*}_{2}$ but not $c^{*}_{k}$'' at the same time.
Then the superposition instance retriever will first encode the query and text into dense representations using a deep neural network model respectively:
\begin{equation}
\mathbf{x}={\rm SIR}(x),\mathbf{q}={\rm SIR}(q)
\end{equation}
After that, the confidence score $s_{x,q}$ indicating the text piece containing a mention of superposition concepts in the set is calculated by:
\begin{equation}
s_{x,q} = \mathbf{x} \circ \mathbf{q}
\end{equation}
where $\circ$ is inner product. We then iteratively select instances of each query with highest confidence scores as the candidate superposition instances to annotate. If the number of queries larger than the annotation budget, we will order the queries based on the frequencies of the excluded concept.

Finally, annotators are asked to annotate the instances retrieved from SIR. Then additional annotated instances and original illustrative instances are used to learn FS-NER models. In this way, the high-value precise generalization knowledge entailed in the additional annotated instances can be injected to the FS-NER models.

\section{Learning CE and SIR}
\begin{table*}[th!]
\centering
\resizebox{0.85\textwidth}{!}{
\begin{tabular}{@{}clcccccc@{}}
\toprule
                        &                                             & Vanilla & Random & ALPS  & BERT-KM & Vote-k    & SuperCD        \\ \midrule
\multirow{5}{*}{WNUT17} & BERT                                        & 27.53   & 29.06  & 26.34 & 28.62   & 27.92          & \textbf{34.16} \\
                        & StructShot~\citep{yang-katiyar-2020-simple} & 30.40   & 33.64  & 33.57 & 34.58   & 33.56          & \textbf{34.74} \\
                        & NSP~\citep{huang-etal-2021-shot}            & 34.20   & 43.67  & 40.15 & 40.69   & 39.97          & \textbf{44.36} \\
                        & CONTaiNER~\citep{das-etal-2022-container}   & 32.50   & 35.23  & 35.23 & 35.14   & 34.81          & \textbf{36.77} \\
                        & SDNet~\citep{chen-etal-2022-shot}           & 44.10   & 45.07  & 44.87 & 44.93   & 43.03          & \textbf{45.63} \\ \midrule
\multirow{5}{*}{WNUT16} & BERT                                        & 26.65   & 35.04  & 31.93 & 30.65   & 34.34          & \textbf{37.22} \\
                        & StructShot~\citep{yang-katiyar-2020-simple} & 31.63   & 32.72  & 33.15 & 34.25   & 33.19          & \textbf{34.80} \\
                        & NSP~\citep{huang-etal-2021-shot}            & 38.40   & 42.26  & 42.06 & 43.14   & 42.02          & \textbf{45.09} \\
                        & CONTaiNER~\citep{das-etal-2022-container}   & 31.07   & 32.30  & 32.53 & 33.80   & 32.97          & \textbf{35.52} \\
                        & SDNet~\citep{chen-etal-2022-shot}           & 47.29   & 49.89  & 49.98 & 50.67   & 49.68          & \textbf{50.72} \\ \midrule
\multirow{5}{*}{CoNLL}  & BERT                                        & 67.88   & 73.86  & 74.20 & 73.26   & 70.54          & \textbf{74.42} \\
                        & StructShot~\citep{yang-katiyar-2020-simple} & 74.80   & 76.63  & 76.74 & 77.22   & \textbf{78.75} & 77.52          \\
                        & NSP~\citep{huang-etal-2021-shot}            & 61.40   & 76.21  & 75.25 & 76.15   & 74.50          & \textbf{77.97} \\
                        & CONTaiNER~\citep{das-etal-2022-container}   & 75.80   & 78.18  & 78.76 & 79.22   & 80.09          & \textbf{80.17} \\
                        & SDNet~\citep{chen-etal-2022-shot}           & 71.40   & 75.88  & 74.88 & 74.99   & 76.15          & \textbf{76.17} \\ \midrule
\multirow{5}{*}{ACE05}  & BERT                                        & 62.66   & 65.77  & 68.19 & 65.84   & 65.06          & \textbf{68.23} \\
                        & StructShot~\citep{yang-katiyar-2020-simple} & 50.63   & 51.40  & 51.84 & 53.37   & 50.83          & \textbf{53.73} \\
                        & NSP~\citep{huang-etal-2021-shot}            & 63.73   & 67.29  & 66.82 & 67.10   & 66.56          & \textbf{68.23} \\
                        & CONTaiNER~\citep{das-etal-2022-container}   & 64.12   & 65.63  & 66.74 & 65.04   & 65.98          & \textbf{67.68} \\
                        & SDNet~\citep{chen-etal-2022-shot}           & 64.78   & 69.43  & 70.99 & 70.03   & 69.44          & \textbf{71.68} \\ \midrule
\multicolumn{2}{c}{AVE}                                               & 50.05   & 53.96  & 53.71 & 53.93   & 53.47          & \textbf{55.74} \\ \bottomrule
\end{tabular}}
\caption{Micro-F1 scores of 5+5-shot FS-NER on test set. Vanilla indicates that the model is trained using the initial illustrative data. The annotated budget is 5 sentences for each type. AVE are the average scores of these datasets and few-shot NER models.}
\label{tab:mainresult}
\end{table*}

To equip Concept Extractor with ability of extracting universal concepts and Superposition Instance Retriever with ability of instance retrieval, we learn them on large-scale, easily accessible Wikipedia and Wikidata. For CE, we follow SDNet~\citep{chen-etal-2022-shot} to learn the ability to extract concepts from texts. For SIR, we train it through a contrastive learning paradigm. 

\subsection{Learning Concept Extractor}
Concept extractor is a sequence-to-sequence model which maps the input instances into their corresponding concepts. In this paper, we leverage the dataset constructed by ~\citet{chen-etal-2022-shot} to train concept extraction. The dataset contains 56M instances in the form of text-to-concepts parallel sequences, which can be directly used to learn the ability of concept extraction.

\begin{figure}[t!]
\centering 
\includegraphics[width=0.3\textwidth]{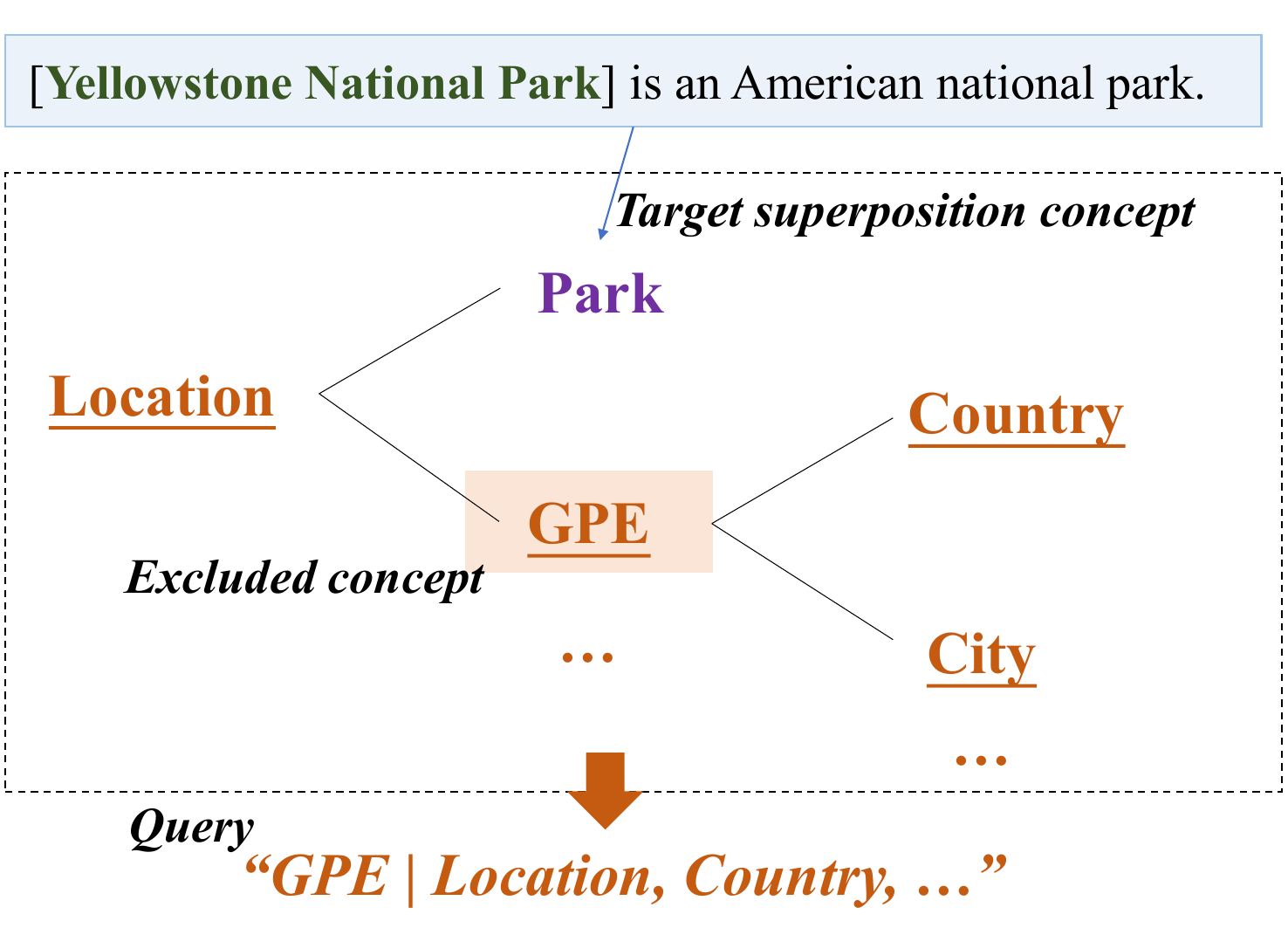}
\caption{An example of query generation in dataset construction.}
\label{Fig:sirdata}
\end{figure}

\subsection{Learning Superposition Instance Retriever}
To equip SIR with the ability of instance retrieval, we construct large-scale query-text pairs from Wikipedia and Wikidata, and leverage a contrastive learning paradigm to learn the retrieval model. Specifically, we regard all sentences in Wikipedia with anchor words to Wikidata items as training instances. We first randomly sampled a corresponding type from a positive instance in Wikidata as the target superposition concept. Then, the excluded concept in the query is selected from the siblings of target concept, and the remaining concepts in the query are sampled from ancestors and descendants of the excluded concept. For example in Figure~\ref{Fig:sirdata}, given an instance ``[Yellowstone National Park] is an American national park.'' and the target concept \textit{Park} is first sampled, we then random choose a sibling concept \textit{GPE} as the excluded concept and the remaining concepts in the query contain \textit{Location}, \textit{Country}, etc. After that, for each query, we will sample two kinds of negative instances, including instances of the excluded concept and randomly sampled instances does not satisfy the query condition. Finally, the dataset contains 10M query and positive instance pairs, where each pair contains about 200 negative instances.

After obtaining the dataset, we propose to use a contrastive learning paradigm to train SIR. Specifically, given a query $q$, a positive instance $x^{(+)}$ and several negative instances $X^-=\{x^{(-)}_1,\ldots,x^{(-)}_N\}$, SIR is learned by optimizing the following loss function:
\begin{equation}\small
L(q,x^+,X^-)=- \log \frac{e^{s_{x^{+},q}}}{e^{s_{x^{+},q}} + \sum_{i=1}^{N}e^{s_{x_i^{-},q}}}
\end{equation}

\section{Experiments}
It is important to note that SuperCD does not aim to achieve state-of-the-art results in FS-NER, but rather to solve the intrinsic precise generalization challenge. Previous FS-NER methods can leverage our approach to improve their performance. In addition, SuperCD is a model-agnostic active learning method that can be universally applied to all types of FS-NER models.

\subsection{Settings}
\paragraph{Active learning setting.}
Unlike previous active learning settings~\citep{shen2018deep,DBLP:journals/corr/abs-2111-03837}, this paper employs an extremely limited budget setting that is consistent with the few-shot goal. Specifically, we assume that the budget is related to the number of target type, and at most $M\times N$ sentences can be annotated (N is the number of target entity types). In a K-shot few-shot setting, active FS-NER will be in a K+M-shot configuration. For each FS-NER dataset, we conduct main experiments on 5+5-shot setting. Following~\citet{huang-etal-2021-shot}, we sample k sentences from the training set to construct initial illustrative instances of each target type in k-shot setting, and the remaining sentences in the training set are used as the unlabeled corpus for active learning. We sample 10 different sets of the illustrative instances. We evaluate the model on the test set with the metric of the average micro-f1 over the 10 runs.

\paragraph{Datasets.} We conducted experiment on 4 few-shot NER datasets with different granularity: 1) WNUT17~\citep{wnut}; 2) ACE2005\footnote{\url{https://www.ldc.upenn.edu/collaborations/past-projects/ace}}, we use the ACE05-E processed by \citet{wadden-etal-2019-entity} and \citet{lin-etal-2020-joint}; 3) CoNLL2003~\citep{conll}; 4) WNUT16 \citep{wnut16}. WNUT17, ACE2005, CoNLL focus on coarse type like \emph{location} and \emph{organization}. WNUT16 focus on fine type like \emph{company}.

\paragraph{Baselines.} To verify the universality of the proposed method, we conduct experiments on five different FS-NER models in three different types: 1) Linear classifier-based FS-NER models, including BERT (base-uncased)~\citep{devlin-etal-2019-bert} and noising supervised pre-trained RoBERTa (NSP)~\citep{huang-etal-2021-shot}; 2) metric-based FS-NER models, including StructShot~\citep{yang-katiyar-2020-simple} and CONTaiNER~\citep{das-etal-2022-container}; 3) generative FS-NER model, SDNet \citep{chen-etal-2022-shot}. For StructShot and CONTaiNER, we pre-train BERT using OntoNotes as described in the original papers. Due to the varied architectures, logits-based active learning methods like Entropy~\citep{DBLP:journals/corr/abs-2111-03837} cannot be directly applied for all these models. Hence, we primarily evaluate model-agnostic universal active learning methods. Additionally, in Section~\ref{sec:logits}, we compare SuperCD with logits-based active learning methods using specific FS-NER models.

We compared SuperCD with universal active learning approaches for all FS-NER models: 1) random sampling (\textbf{Random}), which selects sentences from unlabeled corpus randomly; 2) Diversity sampling \textbf{BERT-KM}~\citep{yuan-etal-2020-cold}, which selects diversity sentences based on feature space; 3) \textbf{ALPS}~\citep{yuan-etal-2020-cold}, which combines the uncertainty and diversity to select sentences. 4) \textbf{Vote-k}~\citep{su2023selective}\footnote{We conduct the fast vote-k since because it can achieve similar performance to vote-k while being more computationally efficient.}, which selects diverse, representative instances for annotation.

\subsection{Main Results}
The experimental results are shown in Table~\ref{tab:mainresult}. We can see that:

1) \textbf{SuperCD can effectively improve the performance of FS-NER by retrieving high-value instances.} Compared with the vanilla models, SuperCD improves the performance by 11.4\%, and compared with random sampling baselines, SuperCD improves the performance by 3.3\%, which indicates that instances retrieved by SuperCD are high-value and can significantly improve performance on FS-NER. 

2) \textbf{Discriminating superposition concepts is helpful to resolve precise generalization challenge.} Compared with best active learning baselines, SuperCD improves the performance by 3.4\%, which shows that precise generalization challenge is not the problem of identifying boundary or diversity cases which is a concern in traditional active learning methods, and therefore other active learning approaches cannot be used to resolve precise generalization challenge directly.

3) \textbf{SuperCD is a universal method for FS-NER.} SuperCD performs well on datasets of varying granularity. Additionally, it is a model-agnostic method that performs well on FS-NER models with different architectures.

\begin{table}[th!]
\centering
\resizebox{0.45\textwidth}{!}{
\begin{tabular}{@{}l|cccc|c@{}}
\toprule
                                                  & \multicolumn{2}{c}{WNUT17}                           & \multicolumn{2}{c|}{WNUT16}     & \multirow{2}{*}{AVE} \\ \cmidrule(r){1-5}
                                                  & BERT           & \multicolumn{1}{c|}{NSP}            & BERT           & NSP            &                      \\ \midrule
FT-BERTKM~\citep{yuan-etal-2020-cold}             & 30.37          & \multicolumn{1}{c|}{39.99}          & 33.26          & 41.58          & 36.30                \\
EnTropy~\citep{DBLP:journals/corr/abs-2111-03837} & 32.04          & \multicolumn{1}{c|}{42.19}          & \textbf{37.34} & 45.00          & 39.14                \\
Badge~\citep{ash2020deep}                         & 31.38          & \multicolumn{1}{c|}{44.33}          & 35.14          & 41.02          & 37.97                \\
CAL~\citep{margatina-etal-2021-active}            & 33.96          & \multicolumn{1}{c|}{39.70}          & 32.23          & 42.61          & 37.13                 \\
SuperCD                                           & \textbf{34.16} & \multicolumn{1}{c|}{\textbf{44.36}} & 37.22          & \textbf{45.09} & \textbf{40.21}                \\ \bottomrule
\end{tabular}}
\caption{The result of logits-based active learning methods. We conduct 5+5-shot setting.}
\label{tab:logits}
\end{table}

\begin{figure}[t!]
\centering 
\setlength{\belowcaptionskip}{-0.4cm}
\includegraphics[width=0.4\textwidth]{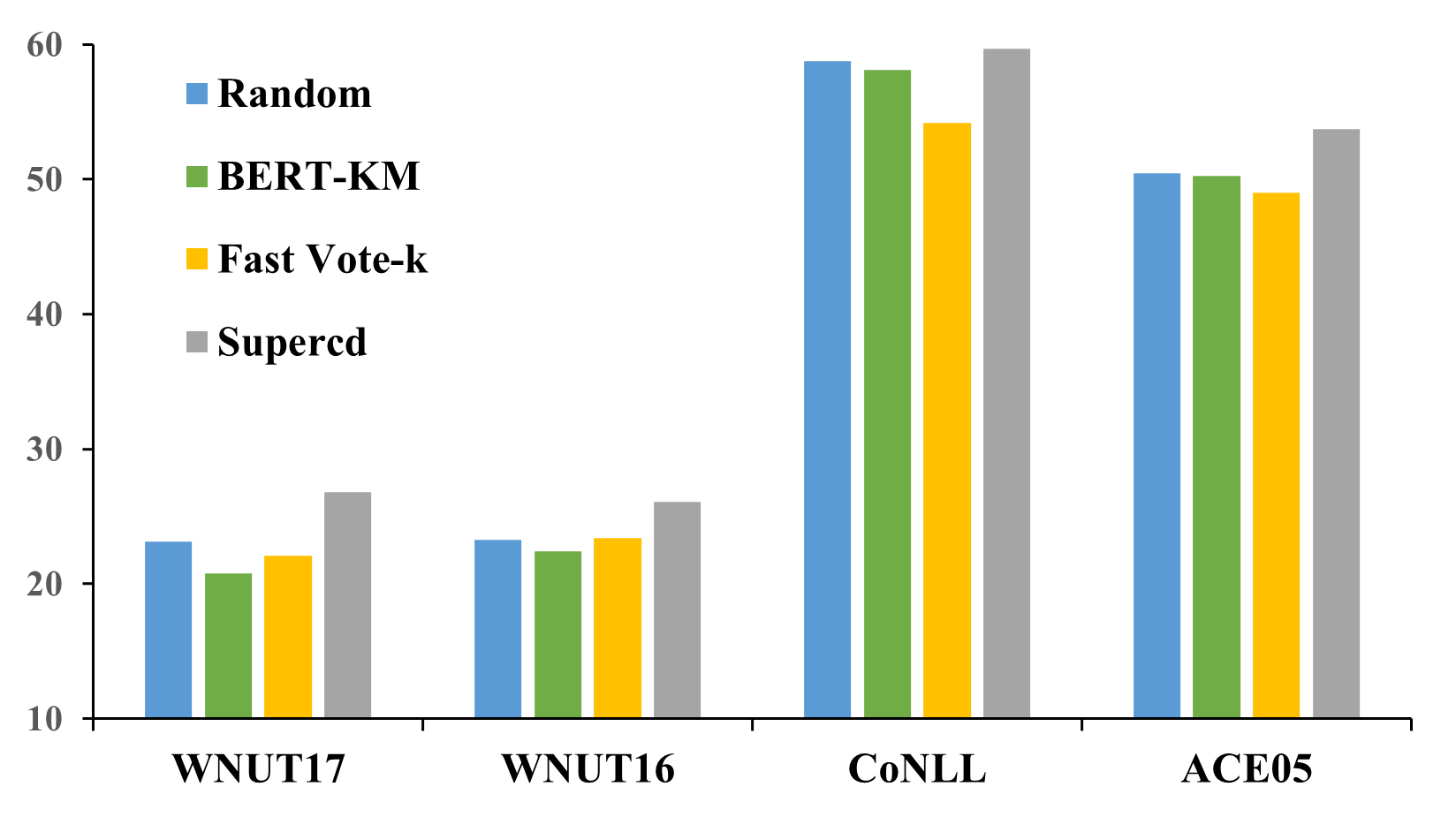}
\caption{The micro-F1 scores of BERT on entities with unseen concepts in the test set. }
\label{Fig:unseen}
\end{figure}

\begin{figure}[t!]
\centering 
\includegraphics[width=0.4\textwidth]{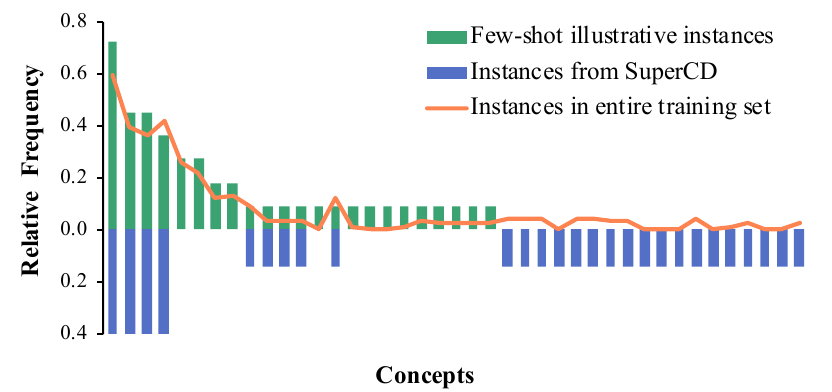}
\caption{The relative frequency distributions of concepts for few-shot illustrative instances, the annotated instances by SuperCD and entire training set instances in \emph{Location} of WNUT17.}
\label{Fig:dist}
\end{figure}

\begin{figure}[t!]
\centering 
\subfloat[WNUT17.]{
		\includegraphics[width=0.23\textwidth]{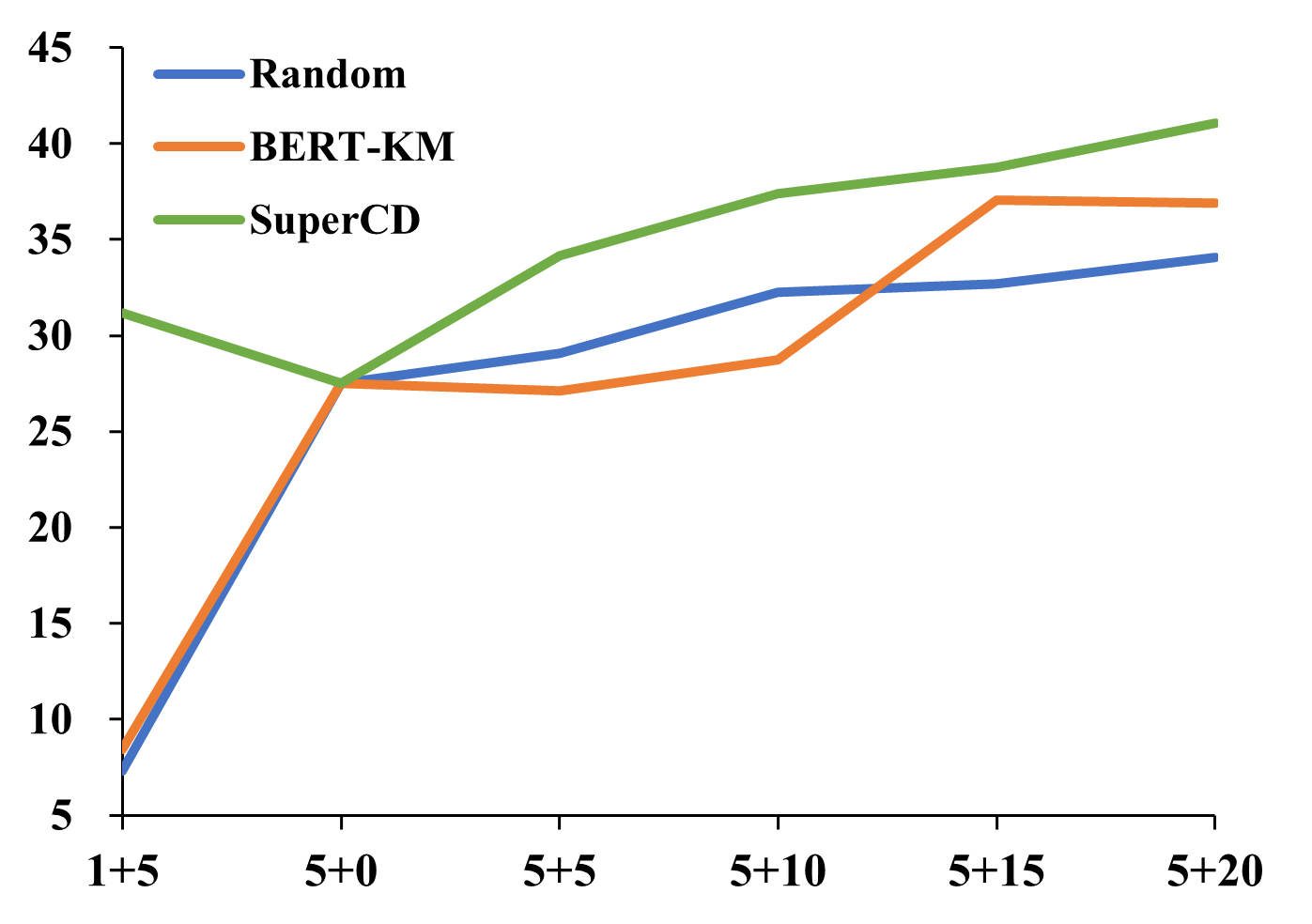}}
\subfloat[WNUT16.]{
		\includegraphics[width=0.23\textwidth]{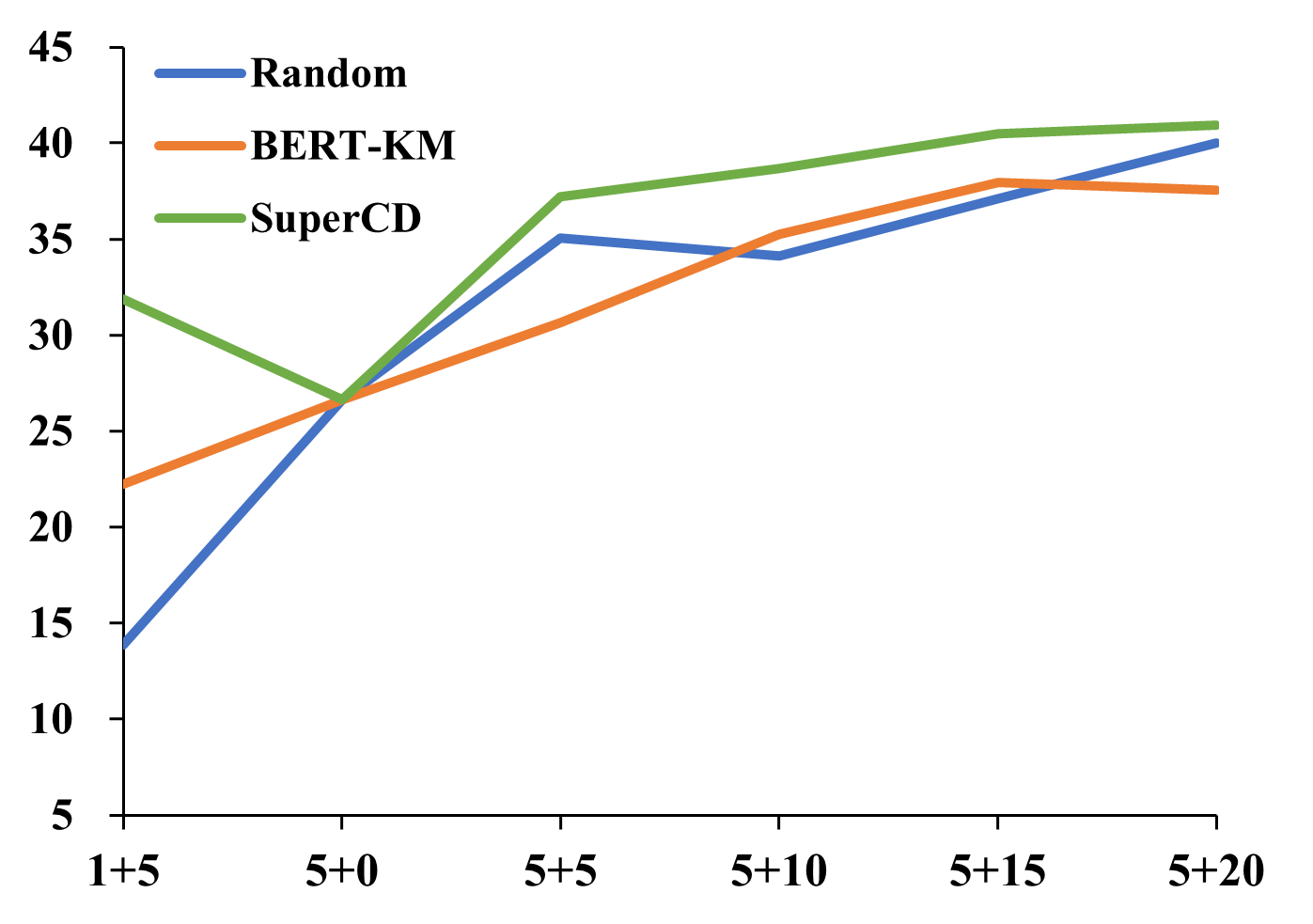}}
\caption{Performances under different annotated budgets. We use \textbf{BERT-KM} to represent the baseline method.}
\label{Fig.k-shot}
\end{figure}

\begin{table*}[]
\centering
\resizebox{0.9\textwidth}{!}{
\begin{tabular}{@{}cll@{}}
\toprule
\multicolumn{3}{c}{\textbf{WNUT17}} \\ \midrule
\multicolumn{1}{c|}{\textbf{Target types}} & \multicolumn{1}{c|}{\textbf{Common concepts of illustrative instances}} & \multicolumn{1}{c}{\textbf{Entities in samples from SuperCD}} \\ \midrule
\multicolumn{1}{c|}{\emph{Location}} & \multicolumn{1}{l|}{\begin{tabular}[c]{@{}l@{}}\emph{\textcolor{teal}{City}},  \emph{\textcolor{teal}{Country}}, \emph{\textcolor{teal}{Human settlement}},\\ \emph{\textcolor{teal}{Educational insititution}}, \emph{\textcolor{teal}{Location}}......\end{tabular}} & \begin{tabular}[c]{@{}l@{}}Glenveagh national park $\to$ \emph{\textcolor{teal}{park}}\\ UNF Arena $\to$ \emph{\textcolor{teal}{arena}}\end{tabular} \\ \midrule
\multicolumn{1}{c|}{\textbf{Test text}} & \multicolumn{2}{l}{…. places such as \textcolor{red}{\textless{}location\textgreater{}}\textbf{Siachen Glacier}\textcolor{red}{\textless{}/location\textgreater} are …} \\ \midrule
\textbf{Vanilla} & \multicolumn{2}{l}{…. places such as Siachen Glacier are provided with the ….} \\
\textbf{SuperCD (ours)} & \multicolumn{2}{l}{…. places such as \textcolor{red}{\textless{}location\textgreater{}}\textbf{Siachen Glacier}\textcolor{red}{\textless{}/location\textgreater} are …} \\ \midrule \midrule
\multicolumn{3}{c}{\textbf{WNUT16}} \\ \midrule
\multicolumn{1}{c|}{\textbf{Target types}} & \multicolumn{1}{c|}{\textbf{Common concepts of illustrative instances}} & \multicolumn{1}{c}{\textbf{Entities in samples vetoed by SuperCD}} \\ \midrule
\multicolumn{1}{c|}{\emph{company}} & \multicolumn{1}{l|}{\begin{tabular}[c]{@{}l@{}}\emph{\textcolor{teal}{Organization}}, \emph{\textcolor{teal}{Company}},  \emph{\textcolor{teal}{Website}}, \\ \emph{\textcolor{teal}{Social networking service}},  \emph{\textcolor{teal}{Business}},......\end{tabular}} & \begin{tabular}[c]{@{}l@{}}NRSC $\to$ \emph{\textcolor{teal}{government agency}}\\ Stanford$\to$ \emph{\textcolor{teal}{university}}\end{tabular} \\ \midrule
\multicolumn{1}{c|}{\textbf{Test text}} & \multicolumn{2}{l}{FBI hack : ex-special agent says those responsible ...} \\ \midrule
\textbf{Vanilla} & \multicolumn{2}{l}{\textcolor{red}{\textless{}company\textgreater}\textbf{FBI}\textcolor{red}{\textless{}/company\textgreater} hack : ex-special agent says those responsible ...} \\
\textbf{SuperCD (ours)} & \multicolumn{2}{l}{FBI hack : ex-special agent says those responsible ...} \\ \bottomrule
\end{tabular}}
\caption{Cases from WNUT17 and WNUT16. For WNUT17, the vanilla BERT is under-generalized and cannot recognize ``Siachen Glacier'' as \emph{Location}. For WNUT16, the vanilla BERT is over-generalized and misidentifies ``FBI'' as \emph{Company}. Note that ``Siachen Glacier'' and ``FBI'' are not in the instances retrieved by SuperCD.}
\label{tab:case}
\end{table*}

\subsection{Comparing SuperCD and logits-based methods.}\label{sec:logits}
To further validate the effectiveness of SuperCD, we conduct logits-based active learning methods as additional baselines: including \textbf{Entropy}~\citep{DBLP:journals/corr/abs-2111-03837},  fine-tuned BERT-KM (\textbf{FT-BERTKM})~\citep{yuan-etal-2020-cold}, \textbf{Badge}~\citep{ash2020deep} and \textbf{CAL}~\citep{margatina-etal-2021-active}. The result is shown in Table~\ref{tab:logits}. We can see that SuperCD can achieve better or competitive performance across datasets of varying granularity. In addition, we found that for coarse-grained dataset WNUT17, SuperCD provides more significant performance improvements than fine-grained dataset WNUT16. This is because fine-grained datasets suffer from a lower risk of under-generalization and the main problem arises from over-generalization. In contrast, coarse-grained datasets suffer from both over-generalization and under-generalization, and therefore SuperCD achieves a more significant improvement.

\subsection{Effectiveness of identifying superposition concepts}\label{sec:overunder}
To further validate the effectiveness of SuperCD, we evaluate the performance of BERT on entities in the test set that contain concepts not seen in the initial illustrative instances. The result is shown in Figure~\ref{Fig:unseen}. We can see that: 1) Compared to the performance on the full test set, entities with unseen concepts perform significantly worse, which reflects the difficulty of the model to learn accurate generalization of the relevant concept by a few illustrative instances, demonstrating the impact of the precise generalization challenge in FS-NER; 2) SuperCD can alleviate the problem by discriminating the superposition concepts. Compare with baselines, SuperCD can significantly improve the performance on entities with unseen concepts, which verifies that retrieved instances play an important role in the performance improvement of FS-NER.

The above conclusions reveal that SuperCD can effectively identify superposition concepts. We further illustrate this by analyzing the relative frequency distributions of concepts in illustrative instances and annotated instances from SuperCD.  We take \emph{location} in WNUT17 as an example and the result is shown in Figure~\ref{Fig:dist}. Since the distribution of entire instances is long-tailed, few-shot illustrative instances contain only a few tailed concepts, which makes it difficult in precise generalization. In contrast, instances retrieved by SuperCD contains many tailed concepts which are high-value superposition concepts and cannot be accessed by simply sampling more instances. This demonstrates that SuperCD can effectively identify superposition concepts.

\subsection{Effectiveness of SuperCD w.r.t. Annotation Budgets}To further validate the effectiveness of SuperCD, we evaluate the performance of BERT under different budget scenarios, i.e., annotating instances from 5 to 20 for each type and we also conduct 1+5-shot setting. The result is shown in Figure~\ref{Fig.k-shot}, we can see that 1) SuperCD works well under different budget scenarios, which indicates that SuperCD is a promising active learning framework for FS-NER. 2) SuperCD outperforms the baseline approach, which indicates that the instances of superposition concepts help FS-NER models determine what the target entity type is.

\section{Case study}
We illustrate the effectiveness of SuperCD by some cases shown in Table~\ref{tab:case}. For coarse-grained dataset WNUT17, when the common concepts of illustrative instances are concepts like \emph{City} and \emph{Country}, vanilla BERT fails to recognize ``Siachen Glacier'' as \emph{Location}. This indicates that the model is under-generalized and has difficulty in determining whether \emph{Glacier} is part of the target type. In contrast, SuperCD discriminates the superposition concepts by annotating some instances of \emph{Park} and \emph{Arena}. In this way, the FS-NER model can understand the scope of the target type better and avoid under-generalization. 

For fine-grained dataset WNUT16, when the target type is \emph{Company}, vanilla BERT misidentifies ``FBI'' as it. Note that the concepts of ``FBI'' contain some common concepts of illustrative instances such as \emph{Organization}. This indicates that the model over-generalize \emph{Company} to \emph{Organization}. In contrast, SuperCD retrievals entity mentions of superposition concepts \emph{University} and \emph{Government agency}. These superposition concepts are vetoed as the corresponding entity mentions are not annotated as target types. By learning from additional instances of superposition concepts, the FS-NER model can understand that the target type is not \emph{Organization} and avoid over-generalization.

\section{Conclusion}
In this paper, we propose Superposition Concept Discriminator for FS-NER  which resolves the precise generalization challenge by requiring annotators to annotate minimal additional high-value instances of superposition concepts. We learn the import components of SuperCD -- Concept Extractor and Superposition Instance Retriever through large-scale, easily accessible web resources. Experiments on 5 datasets show that SuperCD is effective. For future work, we will extend superposition concepts and SuperCD to other few-shot tasks like object detection and event detection.

\section{Limitations}
SuperCD is an active learning algorithm, therefore currently human annotators are still needed, although there are very limited instances to annotate. Furthermore, the performance of SuperCD may be influenced by error propagation from the Concept Extractor and the Superposition Instance Retriever. For instance, the concept coverage of Concept Extractor may influence the performance of SuperCD, which can be improved by introducing more knowledge to learn Concept Extractor.

\section{Acknowledgments}
This work is supported by the Natural Science Foundation of China (No. 62106251, 62122077 and 62306303).

\bibliography{anthology,dataset}

\end{document}